# Lifelong Knowledge Learning in Rule-based Dialogue Systems

**Bing Liu** [1] and **Chuhe Mei** [2]

[1] University of Illinois at Chicago
[2] Peking University
liub@uic.edu

One of the main weaknesses of current chatbots or dialogue systems is that they do not learn online during conversations after they are deployed. This is a major loss of opportunity. Clearly, each human user has a great deal of knowledge about the world that may be useful to others. If a chatbot can learn from their users during chatting, it will greatly expand its knowledge base and serve its users better. This paper proposes to build such a learning capability in a rule-based chatbot so that it can continuously acquire new knowledge in its chatting with users. This work is useful because many deployed chatbots are rule-based.

## 1 Introduction

Chatbots or dialogue systems are very popular in recent years due to a wide range of applications. They often need a large amount of knowledge in order to serve their users well. The knowledge is typically compiled and added offline by engineers. However, the amount of world knowledge in any subject area is too large to be fully covered by a small group of engineers, not to mention that everything changes constantly. Although human users may have a great deal of knowledge of many areas, the current chatbots cannot learn from them. This is a major loss of opportunity. If knowledge can be acquired from users interactively during chatting, the chatbot will become more and more knowledgeable and powerful (Liu and Mazumder, 2021). We humans, in fact, learn a great deal of our knowledge in our daily conversations. This is an instance of *lifelong learning* (Chen and Liu, 2018).

Although deep learning based dialogue models are the mainstream in research, it is also *common knowledge* that most chatbots in deployment are written mainly with handcrafted rules due to high quality, controllable responses, and the fact that rules are relative easy to improve and to update, which are key considerations in building a product. In this paper, we propose to build a knowledge learning capability in a rule-based chatbot through *knowledge distillation patterns* so that the chatbot can perform *continuous* or *lifelong learning* (Chen and Liu, 2018) of new knowledge from users to advance this very useful and practical technology, which is non-less important than designing new deep learning dialogue models, which are still not suitable for many applications. We hope that this learning capability will make its way to the next generation of chatbot products so that they can learn by themselves to become much smarter.

We are aware that there are existing systems (e.g., NELL (Mitchell et al., 2015)) that can sift through Web documents to extract knowledge that can be used in chatbots. By no means do we intend to replace these systems. Our work is, in fact, complementary as we humans learn a great deal of knowledge from books and also from our daily conversations. Clearly, some knowledge acquired from users is inevitably incorrect, but in a multi-user environment, the system can perform cross-verification to verify each piece of learned knowledge by asking other users at appropriate times before the knowledge is stored for future use.

This paper proposes to learn or acquire new knowledge online during interactive chatting using *knowledge distillation patterns*. Our system is called KAD (*knowledge acquisition in dialogues*).

## 2 Related Work

Early chatbots were mainly built with rules and retrieval models (Banchs and Li, 2012; Ameixa et al., 2014; Lowe et al., 2015; Serban et al., 2015). Recently, deep learning models are the mainstream in research (Vinyals and Le, 2015; Xing et al., 2017; Li et al., 2017, Shu et al, 2019). Many of them also use knowledge bases (KB) (e.g., Ghazvininejad et al., 2018; Le et al., 2016; Long et al., 2017; Zhou et al., 2018; Eric and Manning, 2017; Madotto et al., 2018). However, these systems do not learn during dialogue or chatting.

The most closely related works to ours are those in (Mazumder et al., 2019, 2020a), which attempt to build an engine for learning factual knowledge during chatting. Their approach starts with a user query that the system cannot answer. The system then asks for some knowledge from the user in order to infer the answer. Both the user answers and the system inferred query answers are regarded as the acquired knowledge. However, they do not have a dialogue system as they directly use structured triples. Our KAD is a dialogue system and it does not have the restriction of requiring a user query that the system cannot answer. KAD aims to learn from any utterances. Hixon et al. (2015) proposed a system that starts with a multiple-choice science question and based on the multiple answers, it asks the user some questions. KAD is more general and not based on any fixed queries or questions. Mazumder et al. (2020b) also learns to ground language expressions continually.

KAD is related to the work in (Li et al. 2017) and (Zhang et al. 2017) whose goal is to train dialogue systems using human teachers who can ask and answer the system's questions. Our work learns directly from user utterances. The work in (Otsuka et al., 2013; Ono et al., 2017) asks the user whether the system's prediction of category of a term is correct or not. KAD is also remotely related to interactive language learning (e.g., Wang et al., (2016) for knowledge grounding, but this work does not learn any new knowledge like KAD.

## 3 Proposed System

In this work, we focus on knowledge that can be expressed as triples, ($e_1$, $r$, $e_2$), which means that entities $e_1$ and $e_2$ have the relation $r$. For example, (Miami, city-in, USA) means "*Miami is a city in the USA.*" There are many opportunities that can be exploited to learn new knowledge in chatting. This work focuses on only one opportunity, i.e., when a user utterance contains knowledge that may be distilled as a set of triples. This work assumes that all relations are known and are in the *knowledge base* (KB). It thus learns only new instances of the relations, and new entities and their properties.

### 3.1 Learning Using Distillation Patterns

A rule-based dialogue system basically works by matching a user utterance $u$ with a rule in the system and the system then generates a response for the user based on the rule. In the proposed KAD system, a NER (named-entity recognition) system is first applied to identify the entities in $u$ before performing rule matching. Further, in addition to generating a response for the user, the system also learns some new knowledge from $u$ in the background by instantiating *knowledge distillation patterns* (KDPs) attached to the rule with the matching results, and then incorporates the knowledge into the KB. Each instantiated KDP is a piece of knowledge.

We now introduce KDPs, which are attached to rules. Not all rules will be attached with KDPs. Whether a rule should be attached with KDPs depends on if any knowledge can be distilled or extracted from a user utterance that matches the rule. Like rules, KDPs attached to the rules are also written by engineers or the system developer. A rule with attached KDPs is specified as follows:

($p$, $F$, $B$), where $F$ and/or $B$ can be empty.

$F$: a set of KDPs representing fact triples {$f_1$, $f_2$, …, $f_n$} that are true when instantiated with the match results of the rule and the user utterance $u$ and can be added to the KB if they are not already there without confirming with the user.

***Example***: We have the following enhanced rule (a rule with KDPs) for a hotel application:

$p$: [* stayed in X at Y]
$F$: {(X, is-a, hotel), (X, has-address, Y)

If the user said "*I stayed in the Holiday Inn at 150 Pine Street last night.*" This utterance matches the above rule. Two pieces of knowledge (called candidate facts) from $F$ should be true and added to the KB, which are

(Holiday Inn, is-a, hotel) and
(Holiday Inn, has-address, 150 Pine Street).

However, before adding, we need to make sure that they are correct via *cross-verification*.

***Cross-verification***: As we discuss earlier that a piece of knowledge or candidate fact $f_i$ from the user may not be correct or trustworthy, we perform *cross-verification* by asking other users of the system whether it is correct. However, in order to ask about $f_i$, we need to know what question to ask. The question is attached to the relation *is-a* or *has-address* in the KB, which we will discuss in the next sub-section. For the two example facts above in $F$, the cross-verification questions are "*Is X a hotel?*" and "*Is there a X hotel at Y?*" respectively. If the cross-verification questions get affirmative answers from some other users, the above two candidate fact triples can be added to the KB.

For any other answer, the corresponding fact triple (knowledge) is deleted. Note that *Holiday Inn* has been identified by the NER system as a single entity. We allow an entity to have multiple values for a property, e.g., *Holiday Inn* can have multiple locations and thus addresses.

$B$: a set of beliefs ($b_1$, $b_2$, ..., $b_m$} that may or may not be true. The beliefs are used because in some cases, we are not sure whether a triple can be implied by the utterance matching $p$. In this case, for each $b_i$, the system needs to ask the user to confirm its true value. Each $b_i$ consists of two specifications ($m_i$, $A_i$), where $m_i$ is the main belief and $A_i$ is a set of auxiliary facts {$a_1$, $a_2$, ..., $a_k$} that are true and should be added to the KB if $m_i$ is true and is added to the KB. They represent the properties and values about the entities involved in the main belief. $m_i$ is also implicitly associated with two questions: a *confirmation question* to ask the current user and a *cross-verification question* to ask other users to ensure correctness. Again, the questions are attached to the relations in the KB, which we will discuss in the next sub-section.

***Example***: We have the following enhanced rule (a rule with KDPs) for a hotel application:

$p$:  (* stayed in X at Y * with Z friends *)
$B$:  main: (X, is-a, hotel)
     $A$ = {(X, has-address, Y)}

In this case, we use beliefs because we are not sure whether X is a hotel. The user utterance "*I stayed in Holiday Inn at 150 Pine Street last night with a few friends*" matches the above pattern. The system needs to make sure that (Holiday Inn, is-a, hotel) is true (i.e., *Holiday Inn is a hotel*). If Holiday Inn is already in the KB and is a hotel, the system will not ask the user but just add the following to the KB:

(Holiday Inn, has-address, 150 Pine Street)

Otherwise, it will ask the user "*Is Holiday Inn a hotel?*" If the user gives an affirmative answer, the following two pieces of knowledge will be added to the KB if the cross-verification questions receive affirmative answers:

(Holiday Inn, is-a, hotel)
(Holiday Inn, has-address, 150 Pine Street)

If the answer is no, no knowledge will be learned from this utterance.

The newly learned knowledge can be used in future conversations to make the system smarter.

Note that when we say a triple is added to the KB, we mean that we only add the missing information. For example, if the entities are already in the KB, we only add the relation. In order to know whether an entity is already in the system, we need to deal with people using different names to refer to the same entity. We deal with it by simply comparing words using edit distance and then ask the user to confirm using natural language, e.g., "*is Panera the same as Panera Bread?*" For applications where multiple entities have the same name, an existing entity linking system can be used to resolve the confusion and confirm with the user.

### 3.2 Knowledge Base (KB)

We now discuss the KB, which has additional information attached to a traditional KB as a knowledge graph. Our KB consists of a set of triples of the form ($e_1$, $r$, $e_2$) and the triples are organized as a graph, where the nodes $N$ are entities and links are relations $R$. Each entity $N_i$ has an *is-a* type hierarchy and each type (e.g., hotel) has a set of properties (e.g., address, parking, etc.) that will be used by each instantiated instance of the type.

Each relation is attached with a *factual question* $q^f$ that the system asks the current user about a particular relation instance and a *cross-verification question* $q^v$ that the system asks some other users whether the relation instance is true.

We need a special treatment for property relations because we want to ask for property values of a known entity. Needless to say that each entity instance of a type inherits the properties of the type. For example, for the property relation *has-address* for the hotel type, we can have,

$q^f$: *What is the address of X?*
$q^v$: *Is there a X hotel at this address, Y?*

where $X$ is the entity instance, e.g., *Holiday Inn*. With the attached factual question, the system can ask for the property value of a particular instance of an entity type, e.g., "*what is the address* of the *Holiday Inn?*" The system can learn from the answers of the user. Likewise, cross-verification questions can be asked by using $q^v$, e.g., "*Is there a Holiday Inn at this address, 150 Pine Street?*"

For non-property relations, $q^f$ is a confirmation question. For example, for the *is-a* type relation, we may have the following questions ($q^f$ and $q^v$ happen to be the same in this case):

$q^f$: *Is X a hotel?*
$q^v$: *Is X a hotel?*

Note that the system needs to remember the correference when it asks the user and get answers. For example, the current user utterance mentioned

*Holiday Inn*. When the system asks "*what do you like about this hotel*?" or "*which aspects do you like most about the hotel*?" and the user answers "*I like the service*" or "*I love their bed*," each answer may also match a rule, which may imply some other knowledge. However, the system needs to know the hotel that they refer to is *Holiday Inn*.

### 3.3 Asking Questions Later

When an entity is added to the KB, we usually have little information about its properties. Based on the questions discussed above, we can ask the user one by one. However, we cannot ask the same user too many questions as he/she will get irritated. We thus need to keep a queue of all property questions and ask this or other users on the platform later when it is appropriate (this decision is out of the scope of this work). For all kinds of cross-verification questions (property or other relations), we have to ask later as we cannot ask the current user.

It is important to note that a question to be asked when the entity first appears in a user utterance and a question to be used later may need to be different. This is because when the entity is first mentioned, it is known that the entity (e.g., Holiday Inn) refers to a specific entity. However, if we ask later, we need to uniquely identify the specific entity in the question; otherwise we cannot ask, e.g., in the case of a hotel chain. For a chain hotel, we need the specific address to identify a hotel uniquely. For example, we cannot ask "*what is the address of Holiday Inn*?" because there are so many Holiday Inns. Thus, only when the name and the address are known can we ask a question about the hotel, e.g., "*does the Holiday Inn at 150 Pine Street has free parking*?" In designing questions to be attached to each relation in the KB, we need to consider this factor. Hence, we may need two versions of the question for $q^f$ or for $q^v$ if the entity name is not unique. This raises the problem of which property question to ask first. Clearly, we have to ask the questions that uniquely identify the specific entity first, e.g., address of the hotel, because without knowing them, the other questions make no sense.

### 3.4 Reasoning

A set of first-order logical rules can be included in the system about different relations to infer new knowledge from the existing knowledge in the KB.

## 4 Implementation and Evaluation

The proposed KAD system is implemented using the Apache Jena framework, which also provides a SPARQL server named Fuseki for building applications using the SPARQL language. Fuseki supports reasoning by loading inference rules when it starts up. As for mapping natural language sentences to specific SPARQL operations, we use PyAIML, a python interpreter for AIML (AI Markup Language for writing rule-based dialogue systems). The knowledge triples are stored as RDF triplets. For named entity recognition (NER), we use the NER system in (Manning et al., 2014).

The architecture of KAD is given below. When a user talks to KAD, an operation on the KB is generated by the rules. KB will return the response of the operation to the user. If KB needs more information, it will query the Question Database (which stores those factual and cross-verification questions attached to relations) and send questions to the queue. After a question is answered, KB will incorporate it as a new piece of knowledge.

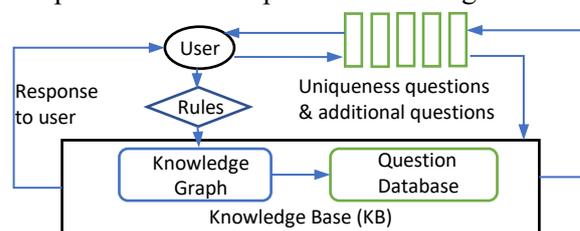

We used KAD in two applications, *restaurants* and *hotels*. A person not involved in the project was asked to collect 100 sentences from the Web for each application. We have rules to distill knowledge from them. Since the rules are manually written, there is no error in knowledge acquisition or in using the knowledge because this work assumes that the relations are known. The system only acquires instance level knowledge.

## 5 Conclusion

We believe that to advance the practical chatbot technology, one key direction is to make chatbots learn from users online during chatting. This paper proposed a pattern-based approach called KAD to achieve this goal. To our knowledge, KAD is the first system with this capability. Although there are some related works, they are very different from KAD as we discussed in the related work.

Clearly, a great deal of work remains to be done, such as how to deal with conflicting knowledge, how to retract a piece of wrong knowledge, how to deal with changing knowledge, and how to learn new relations. Our work in this paper represents a major step towards building a learning chatbot. Our future work will deal with the above problems.